\documentclass[conference,compsoc]{IEEEtran}
\IEEEoverridecommandlockouts
\ifCLASSOPTIONcompsoc
  \usepackage[nocompress]{cite}
\else
  \usepackage{cite}
\fi

\usepackage{amsmath,amssymb,amsfonts}
 
\usepackage{graphicx}
 
\usepackage{booktabs}
\usepackage{textcomp}
\usepackage{xcolor} 
\def\BibTeX{{\rm B\kern-.05em{\sc i\kern-.025em b}\kern-.08em T\kern-.1667em\lower.7ex\hbox{E}\kern-.125emX}}

\begin{document}
 
\title{From Single-Shot QA to Continuous Perception: Revisiting the Energy Composition and Optimization Boundary of Edge VLMs}
 
\author{
\IEEEauthorblockN{Yiting Duan}
\IEEEauthorblockA{\textit{High School Student}\\
\textit{Shanghai Yucai High School}\\
Shanghai, China\\
3397915795@qq.com}
\and
\IEEEauthorblockN{Miaoxin Wang}
\IEEEauthorblockA{\textit{Shanghai Jiao Tong University}\\
Shanghai, China\\
wangmiaoxin26@sjtu.edu.cn}
\and
\IEEEauthorblockN{Aoqing Guo}
\IEEEauthorblockA{\textit{Shanghai Jiao Tong University}\\
Shanghai, China\\
superdtcr@sjtu.edu.cn}
\and
\IEEEauthorblockN{Bo Peng$^{\ast}$}
\IEEEauthorblockA{\textit{Shanghai Jiao Tong University}\\
Shanghai, China\\
pengbo\_michael@sjtu.edu.cn\\
$^{\ast}$Corresponding author}
}

\maketitle

\begin{abstract}
Visual-prefill energy conclusions for edge vision-language models (VLMs) are commonly drawn under single-shot workloads; we revisit them under continuous perception. Profiling nine VLMs across three architecture families on Jetson AGX Thor and Orin (4,508 records, calibrated with a pure-text task), we find two mutually independent energy laws: the per-output-token decode cost grows monotonically with model scale, whereas the visual-prefill (image-encoding) cost is dominated by visual-encoder architectural efficiency, not scale. Under continuous perception ($N{=}10$) with sparse output, visual-prefill energy reaches 92--97\% of total. We derive optimization strategies---visual frame reuse, adaptive output truncation, task-aware routing, and encoder throughput optimization---for edge VLM deployment.
\end{abstract}
\def\IEEEkeywordsname{Keywords}

\section{Introduction}

As large models proliferate, energy has become a primary concern across the computing stack~\cite{Patterson2022,dodge2022carbon}. Battery-powered edge systems---drones, mobile robots, sensors---operate on watt-hour budgets, where a single inference shortens device lifetime~\cite{han2016deep}. Embodied systems are especially constrained: Isaac GR00T processes streaming camera video on Jetson Thor~\cite{gr00t}, and vision-language-action (VLA) models such as $\pi_{0.6}$~\cite{pi06} run on-board. These systems perceive the world continuously: cameras stream frames at 10--30~fps, and the model must encode every frame before deciding whether to generate a response or act. The energy profile of such continuous-perception workloads differs fundamentally from the single-shot question-answering setting commonly studied in the literature, in which a single image is encoded once before a long response is generated.

Zhan et al.~\cite{seeing} were the first to systematically profile the energy consumption of edge VLM inference. Across five VLMs and two platforms, they found that (1) average power is an intrinsic constant of the model (CV$<$5\%); (2) each output token costs 11--39$\times$ more energy than each input token; (3) decoding accounts for 86--97\% of total energy; and (4) total energy is linearly correlated with the number of output tokens. Their headline conclusion follows: image encoding (prefill) is both faster and cheaper than autoregressive decoding, making output-length control the dominant energy-optimization lever.

However, this conclusion rests on two assumptions---that decode cost is predictable from model scale and output length, and that visual-prefill cost is negligible. We profile nine VLMs with two engines on Jetson Thor and Orin (4,508 records, calibrated with a pure-text model) and show that both assumptions are replaced by two independent laws: decode cost per token grows monotonically with scale, while visual-prefill cost is dominated by encoder efficiency.

Our contributions are as follows: (1) we reproduce the four core findings of~\cite{seeing} with eight models on Thor; (2) we reveal the two energy laws---visual-prefill cost is dominated by encoder throughput, which varies 16$\times$; (3) a device-mapping ablation shows that forcing the encoder onto the CPU raises energy by about 27$\times$ and latency by about 37$\times$; (4) cross-engine and cross-machine checks confirm both laws; and (5) we formalize the frame-to-answer crossover $N^{*}$ and derive Joule-priced optimization strategies.

\section{Related Work}

Energy profiling of edge VLM inference was pioneered by Zhan et al.~\cite{seeing}. We reproduce their methodology on newer hardware (Jetson Thor) and a broader model set (nine vs.\ five), then identify where their conclusions break down. Prior LLM energy work~\cite{Patterson2022,dodge2022carbon} focused on data-center scale and carbon accounting; edge studies~\cite{han2016deep} targeted compression and acceleration rather than phase-level energy decomposition.

On-board VLA inference is an active bottleneck: LiteVLA-Edge achieves a 6.6~Hz closed loop through 4-bit GGUF quantization~\cite{litevla}. This work highlights the need for fine-grained energy analysis but does not decompose prefill and decode. Component-level quantization~\cite{vlmquant} shows the visual encoder and language backbone exhibit different sensitivity profiles, confirming that encoder placement is a first-order design parameter. Gemma-3 and Qwen2.5-VL are deployed VLA backbones~\cite{opentau,actionslang} running on Jetson Thor~\cite{gr00t}, making our findings directly relevant to real embodied systems.

\section{Experimental Methodology}

\subsection{Justification of the Joule as the Energy Metric}
The joule (J) is the SI unit of energy ($1~\mathrm{J}=1~\mathrm{W}\cdot\mathrm{s}$) and converts directly to battery watt-hours ($1~\mathrm{Wh}=3600~\mathrm{J}$): a single 943~J inference (gemma-3-12b, 256 tokens) consumes 0.262~Wh, so a 50~Wh battery supports at most 191 such queries. Critically, measuring the total board-level input power (the VIN rail on Jetson) captures the energy consumed by all compute units: during gemma's visual-prefill stage both the GPU rail (VDD\_GPU) and the CPU rail (VDD\_CPU\_SOC) are active (29.7 vs.\ 8.1~W), so measuring either rail alone would miss the other side, whereas the board-level definition covers both. This definition is aligned with~\cite{seeing} (HWiNFO64 system power on RTX 3070 and INA3221 VDD\_IN on Orin NX), ensuring cross-study comparability. We acknowledge that board-level energy includes static idle power (cancelled as a constant offset in cross-model comparisons) and that we measure total electrical energy rather than useful computational work.

\subsection{Platforms and Protocol}
Primary platform: NVIDIA Jetson AGX Thor (Blackwell, compute capability 11.0), 128~GB unified memory, Ubuntu 24.04, 120~W power mode. Secondary platform: Jetson AGX Orin (Ampere, compute capability 8.7), 50~W power mode, used for cross-machine validation.

Engines: (1) llama.cpp build 10396 with all layers offloaded (\texttt{-ngl 999}), context 4096, greedy decoding (temperature 0), \texttt{--no-cache-prompt} enabled to prevent KV-cache contamination of prefill measurements, and GGUF Q4\_K\_M quantization with an F16 multimodal projector; (2) vLLM with HuggingFace fp16 weights, used for the cross-engine comparison of qwen2.5-vl-3b. Models: nine VLMs covering three architecture families---InternVL3-1B/2B/8B, Qwen2-VL-2B, Qwen2.5-VL-3B, Qwen3-VL-8B, MiniCPM-V2.6 (8B), and gemma-3-4B/12B. gemma-3-12B data come from earlier three-tier output-length experiments and are used mainly for scaling analysis. Protocol: 40 images from the COCO 2017 validation set, stratified by annotated object count into six complexity tiers to control for visual-content variation, $\times$ two question types (long: ``Describe this image.''; short: single-word answers) $\times$ 5 repetitions $=$ 400 rows per model. Each condition is preceded by a warm-up and followed by a 10~s cooldown to prevent thermal throttling. Of the 4,508 total records, 3,554 (79\%) are directly measured; short requests (mainly Qwen2-VL-2B, about 0.2~s) reuse calibrated average power.

\subsection{Measurement and Calibration}
VLM inference is split into prefill (visual prefill) and decode (autoregressive decoding). Phase-level wall-clock times are taken from llama-server timings (\texttt{prompt\_ms}, \texttt{predicted\_ms}), consistent with~\cite{seeing}. Onboard input power is sampled by tegrastats at 100~ms, parsing Thor's proprietary \texttt{VIN} field. Energy is decomposed into visual-prefill and decode components: the visual-prefill part is
\begin{equation}\label{eq:esee}
E_{\mathrm{see}} = \bar{P}\,t_{\mathrm{prefill}},
\end{equation}
and the decode part $E_{\mathrm{tok}}$ is obtained as the slope of the linear fit of $E_{\mathrm{total}}$ versus $N_{\mathrm{out}}$ (the intercept is $E_{\mathrm{see}}$),
\begin{equation}\label{eq:etotal}
E_{\mathrm{total}} = E_{\mathrm{see}} + E_{\mathrm{tok}}\,N_{\mathrm{out}}.
\end{equation}
where $\bar{P}$ is the average onboard power (W), $t_{\mathrm{prefill}}$ is the prefill time (s), and $E_{\mathrm{see}}$, $E_{\mathrm{tok}}$, $E_{\mathrm{total}}$ are the visual-prefill energy (J), per-output-token energy (J/tok), and total energy (J), respectively; $N_{\mathrm{out}}$ and $N_{\mathrm{in}}$ are the numbers of output and input tokens.

We validate the entire pipeline with a pure-text counting task on Qwen3-4B, whose output-token count is precisely controllable and free of visual-encoding confounds. Four checks: (1) power CV$=$1.07\%, below~\cite{seeing}'s 5\% threshold; (2) $E = 1.3205\,N_{\mathrm{out}} - 1.16$ with $R^2{=}0.9999$; (3) decode throughput CV$=$0.9\%; and (4) a slope self-consistency error of 1.9\% (the fitted slope and $\bar{P}/\mathrm{tok\_rate}$ agree within 1.9\%). Recovering this known law confirms that the 100~ms tegrastats power pipeline is trustworthy and that the energy metrics $E_{\mathrm{tok}}$ and $E_{\mathrm{see}}$ are reliable. Additional validation: (i) anti-caching---\texttt{--no-cache-prompt} is enabled for all batches, with $N_{\mathrm{in}}{\approx}277$--360 verified, confirming that prefill is not bypassed; (ii) component power-allocation probe---during the one\_word stage, gemma's VDD\_GPU rail is 29.7~W and the VDD\_CPU\_SOC rail is 8.1~W (27.6/11.4~W for Qwen2.5-VL), confirming that the visual encoder executes on the GPU by default rather than falling back to the CPU (the device-mapping ablation is in Section~\ref{sec:ablation}); and (iii) transformer prefill baseline---gemma-3-4b performs prefill in only 122~ms on a pure-text prompt (11 tokens), showing that the language transformer runs on the GPU; the additional $\sim$1.4~s for image prompts stems from low visual-encoding throughput (182~tok/s), not from device misplacement.

\section{Experimental Results}

\subsection{Reproduction of Reference Findings}\label{sec:repro}
\begin{figure}[!t]
\centering
\includegraphics[width=0.75\columnwidth]{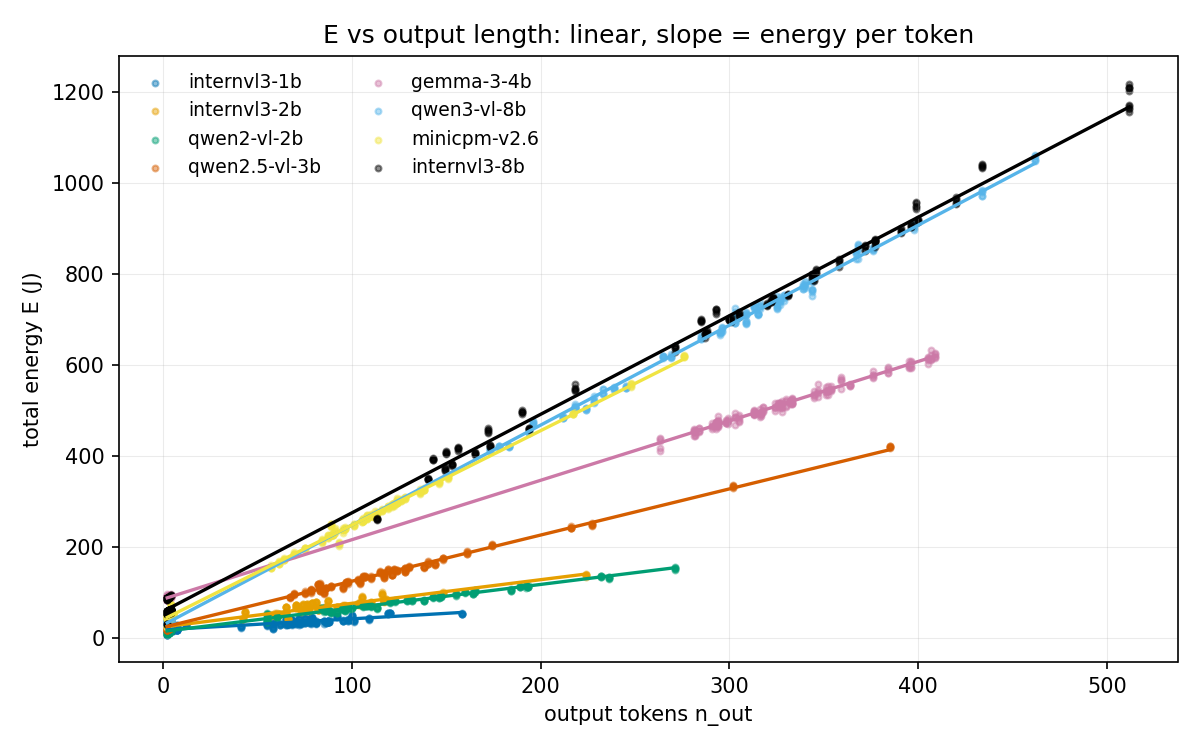}
\caption{Total energy versus the number of output tokens for the eight main-experiment models on Thor.}
\label{fig:erepro}
\end{figure}

Table~\ref{tab:repro} summarizes the reproduction results on Jetson Thor with llama.cpp. We confirm the four core findings of~\cite{seeing}: (1) power CV is below 5\% for all eight models (1.3--4.7\%); (2) $E\sim N_{\mathrm{out}}$ reaches $R^2{\geq}0.969$ for every model, with seven exceeding $R^2{=}0.99$ (Fig.~\ref{fig:erepro}); (3) decoding accounts for 53--82\% of total energy at the describe length (82--331 tokens), lower than the 86--97\% reported in~\cite{seeing}; this discrepancy stems mainly from our broader coverage of visual-encoder architectures (some models incur very high visual-prefill costs) and from 4-bit quantization reducing decode memory access; despite the difference in absolute shares, the qualitative law that ``decode energy scales with output length and dominates it'' still holds; and (4) output tokens cost 11--39$\times$ more than input tokens. The fitted slopes recover each model's $E_{\mathrm{tok}}$, and the intercepts recover $E_{\mathrm{see}}$. We also find that image complexity has no significant effect on prefill time or decode energy for any model---consistent with~\cite{seeing}---confirming that stratified sampling introduces no confound.

\begin{table}[!t]
\centering
\caption{Reproduction on nine VLMs with Jetson Thor (llama.cpp, GGUF Q4\_K\_M). $E_{\mathrm{see}}$ is taken from the short-question row; $J/\mathrm{tok}$ is the per-image-token visual-prefill cost, i.e., the visual-prefill side of the two laws.}
\label{tab:repro}
\footnotesize
\setlength{\tabcolsep}{2.5pt}
\begin{tabular}{lccccc}
\toprule
Model & $S(\mathrm{B})$ & $E_{\mathrm{tok}}$ & $E_{\mathrm{see}}(\mathrm{J})$ & $N_{\mathrm{in}}$ & $J/\mathrm{tok}$ \\
\midrule
internvl3-1b & 1 & 0.28 & 17.7 & 927 & 0.019 \\
internvl3-2b & 2 & 0.50 & 24.6 & 941 & 0.026 \\
qwen2-vl-2b & 2 & 0.51 & 14.6 & 367 & 0.040 \\
qwen2.5-vl-3b & 3 & 1.02 & 25.5 & 367 & 0.070 \\
gemma-3-4b & 4 & 1.31 & 87.9 & 281 & 0.313 \\
qwen3-vl-8b & 8 & 2.22 & 33.7 & 277 & 0.121 \\
minicpm-v2.6 & 8 & 2.10 & 46.9 & 231 & 0.203 \\
internvl3-8b & 8 & 2.19 & 64.6 & 941 & 0.069 \\
gemma-3-12b & 12 & 3.22 & 118.5 & --- & --- \\
\bottomrule
\end{tabular}

{\footnotesize Note: gemma-3-12b used early three-tier output-length data; it conforms to the unified test protocol and is included in the main table.}
\end{table}

\subsection{Decode Energy Is Predictable from Model Scale}

\begin{figure}[!t]
\centering
\includegraphics[width=0.75\columnwidth]{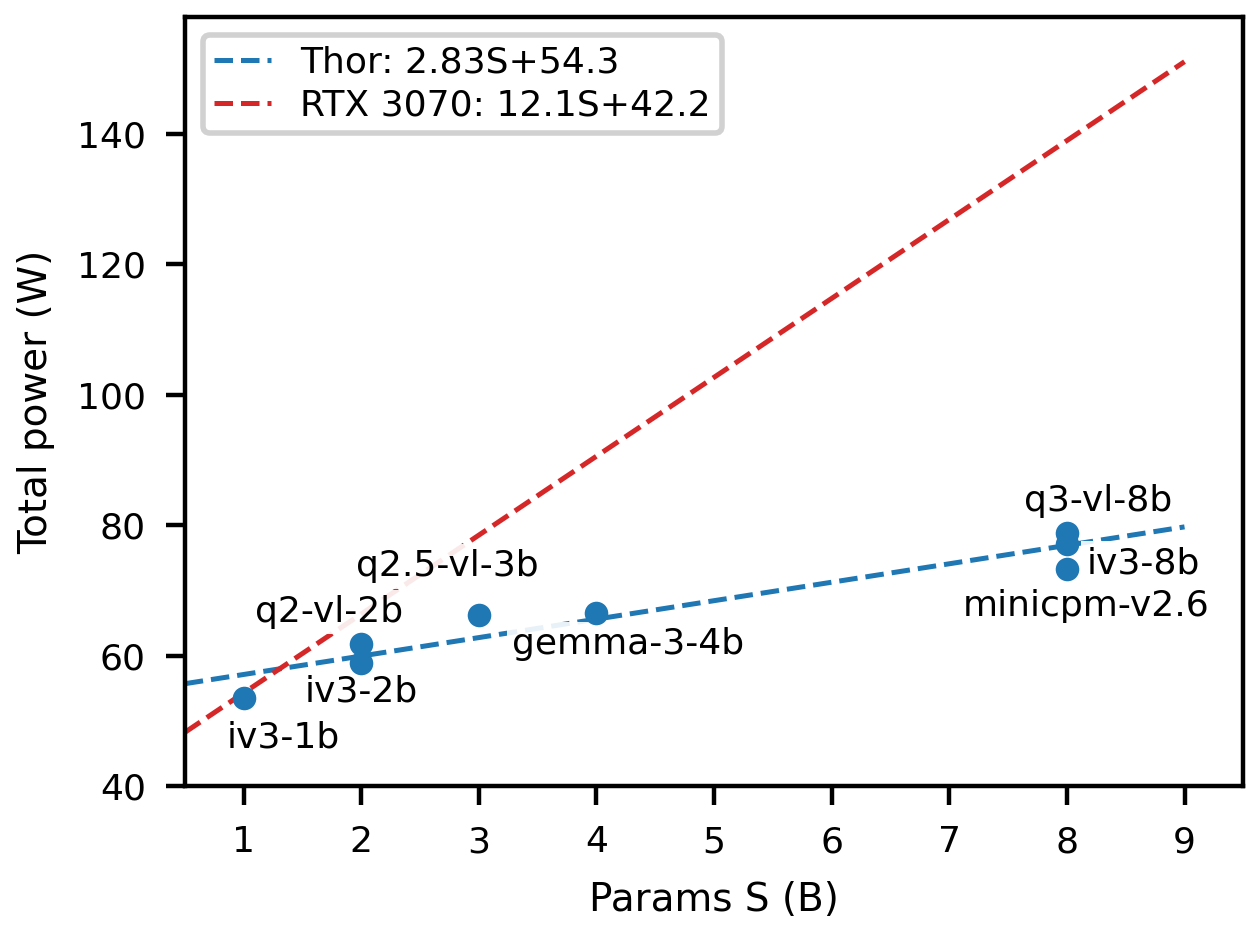}
\caption{Average onboard power versus model parameter count.}
\label{fig:powerscale}
\end{figure}

\begin{figure}[!t]
\centering
\includegraphics[width=\columnwidth]{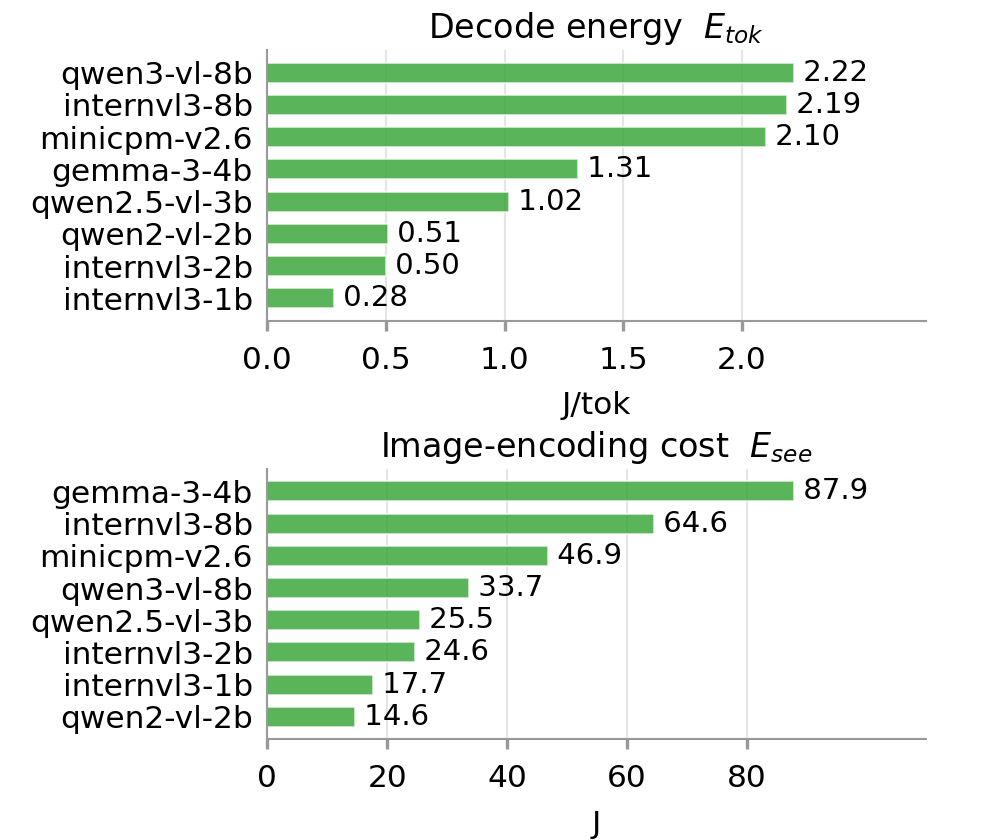}
\caption{Top: the decoding energy law. Bottom: the visual-prefill energy law.}
\label{fig:allbars}
\end{figure}

The per-output-token decoding cost $E_{\mathrm{tok}}$ is the most predictable component of edge VLM energy: over the eight main-experiment Thor models (gemma-3-12b is excluded from the correlation because it used early experimental data), $E_{\mathrm{tok}}$ has a Spearman correlation of $r{=}0.970$ ($p{<}0.001$) with scale $S$ (in billions of parameters), rising monotonically from 0.28~J/tok at 1B to 2.10--2.22~J/tok at 8B, and reaching 3.22 for gemma at 12B---a span of roughly 12$\times$ (Fig.~\ref{fig:allbars}, top). Combined with the linear fits of $R^2{\geq}0.969$ (Section~\ref{sec:repro}), the decode energy $E_{\mathrm{decode}}$ can be budgeted from model scale and output length alone:
\begin{equation}\label{eq:edecode}
E_{\mathrm{decode}} \approx E_{\mathrm{tok}}(S)\cdot N_{\mathrm{out}},
\end{equation}
without any knowledge of visual-prefill-cost details. The power--scale relationship is weaker on Thor: the fit is $\bar{P}_{\mathrm{Thor}} = 2.83\,S + 54.3$ ($R^2{=}0.914$, Fig.~\ref{fig:powerscale}), with a slope far below the reference study's~\cite{seeing} $\bar{P}_{\mathrm{3070}} = 12.1\,S + 42.2$---on Thor the fixed baseline power dominates, and model scale has only a small marginal effect on power, so the scaling signal in decode cost comes mainly from per-token time rather than power. $E_{\mathrm{tok}}$ thus constitutes a predictable ``decoding energy law'': it is a function of model scale, independent of image content.

\subsection{Visual-Prefill Energy Is Dominated by the Encoder Architecture}
In contrast to the decoding law, the visual-prefill cost $E_{\mathrm{see}}$ does not vary monotonically with model scale. Although $E_{\mathrm{see}}$ shows some statistical correlation with scale (Spearman $r{=}0.786$, $p{=}0.021$), model scale is not the dominant factor in the visual-prefill cost. Clear counterexamples to scale monotonicity exist in the data: the 4B-parameter gemma-3-4b incurs a visual-prefill energy as high as 87.9~J, far exceeding the 8B-parameter internvl3-8b (64.6~J); among the three 8B models, the visual-prefill energy spans 1.9$\times$ (33.7--64.6~J). This indicates that cross-architectural differences in visual-encoder implementation efficiency, far beyond the marginal effect of model scale itself, are the core variable determining the visual-prefill cost. Figure~\ref{fig:allbars} (bottom) and Table~\ref{tab:repro} show this disorder. Normalized by image tokens, the per-image-token visual-prefill cost $J/\mathrm{tok}$ varies 16$\times$ across architectures (0.019--0.313~J/tok). The decisive evidence is gemma-3-4b: it has only 281 image tokens (less than one-third of internvl3's 941) yet has the highest per-token cost (0.313~J/tok, 1.5$\times$ the runner-up minicpm's 0.203) and the lowest prefill throughput (182~tok/s, one-sixteenth of internvl3-1b). The mechanism is explained by the energy formula:
\begin{equation}\label{eq:jtok}
\mathrm{J/tok} = \frac{\bar{P}_{\mathrm{prefill}}}{T_{\mathrm{prefill}}};
\end{equation}
onboard power $\bar{P}_{\mathrm{prefill}}$ varies only 1.2$\times$ across models (53.3--66.1~W, one\_word), whereas encoder throughput $T_{\mathrm{prefill}}$ varies 16$\times$ (182--2909~tok/s); throughput is therefore the dominant variable in the visual-prefill cost. Checked against measurements: gemma $55.7/182{=}0.31$ and internvl3-1b $53.7/2909{=}0.019$, in agreement with Table~\ref{tab:repro}. $E_{\mathrm{see}}$ thus constitutes a ``visual-prefill energy law'' that is independent of the decoding law: it is determined by the throughput efficiency of the visual encoder (SigLIP, ViT, etc.) and cannot be predicted from model scale or image-token count.

\subsection{Device-Mapping Ablation: Misplacing the Encoder Massively Increases Energy}\label{sec:ablation}
To separate ``architectural inefficiency'' from ``device misplacement,'' we perform a device-mapping ablation on the same gemma-3-4b: llama.cpp default (visual encoder resident on the GPU, VDD\_GPU 29.7~W) versus forcing CPU execution (\texttt{--no-mmproj-offload}). Under the same image and single-word question, the default prefill is 1.54~s ($E_{\mathrm{see}}{=}87.9$~J); forcing CPU raises prefill to 56.9~s and the visual-prefill energy to about 2{,}370~J (an average of 41.6~W during the active period over a 20~W idle baseline)---an approximately 37$\times$ increase in latency and 27$\times$ in energy. We therefore conclude that the encoder's executing device and implementation efficiency are first-order determinants of the visual-prefill cost, and that device-mapping errors can massively increase energy consumption.

\subsection{Cross-Engine and Cross-Machine Validation}\label{sec:cross}
In the cross-engine experiment (qwen2.5-vl-3b on Thor), llama.cpp (GGUF Q4) yields a per-image-token visual-prefill energy of 0.070~J and $E_{\mathrm{see}}{=}25.5$~J, while vLLM (HF fp16) yields 0.022~J and $E_{\mathrm{see}}{=}8.1$~J (a 3.1$\times$ difference). The engine is thus part of encoder implementation efficiency, again confirming that ``visual-prefill cost depends on implementation, not scale.'' On the decode side the opposite holds: llama.cpp is cheaper (1.02 vs.\ 2.08~J/tok, 2$\times$) because vLLM decodes more slowly (29.1 vs.\ 66.3~tok/s). The two engines exhibit opposite visual-prefill/decoding trade-offs: vLLM is more energy-efficient for short-answer workloads, whereas llama.cpp is more efficient for long-answer workloads.

\begin{table}[!t]
\centering
\caption{Cross-engine validation (Thor, qwen2.5-vl-3b).}
\label{tab:crossengine}
\resizebox{\columnwidth}{!}{%
\begin{tabular}{lccccc}
\toprule
Engine (quantization) & Visual-prefill per token (J/tok) & $E_{\mathrm{see}}$ (J) & Decode $E_{\mathrm{tok}}$ (J/tok) & Decode speed (tok/s) \\
\midrule
llama.cpp (GGUF Q4) & 0.070 & 25.5 & 1.02 & 66.3 \\
vLLM (HF fp16) & 0.022 & 8.1 & 2.08 & 29.1 \\
\bottomrule
\end{tabular}}
\end{table}

Cross-machine (llama.cpp, Thor 120~W vs.\ Orin 50~W): gemma-3-4b's $E_{\mathrm{see}}$ is 3.4$\times$ that of qwen2.5-vl-3b on both machines (Thor 87.9/25.5, Orin 71.2/20.9), and the per-token cost ratio is also reproduced (4.5$\times$ on Thor, 4.4$\times$ on Orin; the two models have identical image-token counts on both machines, so the per-token ratio equals the total-energy ratio), confirming that the visual-prefill energy law is not Thor-specific. Absolute energy values do differ substantially, however: Thor has 2.3--2.4$\times$ higher onboard power and 1.8--1.9$\times$ faster decode throughput, yet per-token energy is only 1.2--1.3$\times$ higher (the higher power is partially offset by the faster speed). Findings B and C therefore hold regardless of hardware.

\begin{table}[!t]
\centering
\caption{Cross-machine validation (llama.cpp, Thor 120~W vs.\ Orin 50~W).}
\label{tab:crossmachine}
\resizebox{\columnwidth}{!}{%
\begin{tabular}{lcccc}
\toprule
Model & Thor $E_{\mathrm{see}}(\mathrm{J})$ & Thor $J/\mathrm{tok}$ & Orin $E_{\mathrm{see}}(\mathrm{J})$ & Orin $J/\mathrm{tok}$ \\
\midrule
gemma-3-4b & 87.9 & 0.313 & 71.2 & 0.253 \\
qwen2.5-vl-3b & 25.5 & 0.070 & 20.9 & 0.057 \\
\bottomrule
\end{tabular}}
\end{table}

\subsection{Continuous Perception Makes Visual-Prefill Energy Dominant}
Real embodied systems have a load profile fundamentally different from single-shot QA: their cameras stream frames continuously, the model re-encodes every frame, and control instructions are usually produced with extremely short token outputs (``high-frequency perception, sparse output''). We model this scenario as encoding $N$ frames before generating one $L$-token response, with total energy
\begin{equation}\label{eq:total}
E = N\,E_{\mathrm{see}} + L\,E_{\mathrm{tok}}.
\end{equation}
Given the typical short-command outputs in embodied control (let $L{=}10$ tokens), when the number of continuously perceived frames is $N{=}10$ (about one second at 10~fps), the visual-prefill energy share reaches 92--97\% for every model; at $N{=}30$, this share further exceeds 97\%. Even qwen2.5-vl-3b, whose visual-prefill energy is tiny in the single-shot setting, is dominated by the visual-prefill stage under continuous perception. We define the frame-to-answer crossover point
\begin{equation}\label{eq:nstar}
N^{*}(L) = \frac{L\,E_{\mathrm{tok}}}{E_{\mathrm{see}}},
\end{equation}
as the number of frames at which the visual-prefill energy exceeds the decode energy. At $L{=}200$ tokens, $N^{*}{=}5.4$ (gemma-12b), 3.0 (gemma-4b), and 8.0 (Qwen-3b)---all under one second at 10~fps. For shorter responses ($L{=}20$), $N^{*}$ drops to 0.5/0.3/0.8, and visual prefill dominates within a single frame. $N^{*}$ generalizes~\cite{seeing}'s output-token crossover to the frame dimension and is a key design metric for perception-driven embodied systems: under the workloads that dominate real robot deployments---continuous perception with sparse output---cost is governed by the visual-prefill energy law ($E_{\mathrm{see}}$) rather than the decoding energy law.

\section{Energy Optimization}\label{sec:opt}
Findings B and C let the benefit of each strategy be quantified in joules: (1) \emph{visual frame reuse}---adjacent near-identical frames reuse the previous frame's image tokens, saving $E_{\mathrm{see}}$ per skipped frame (87.9~J for gemma-4b, 25.5~J for Qwen-3b); reusing 9 of 10 frames eliminates 90\% of the visual-prefill cost. (2) \emph{adaptive output-length control}---makes $E_{\mathrm{tok}}{\cdot}N_{\mathrm{out}}$ linearly controllable; truncating from 256 to 64 tokens saves about 618~J (gemma-12b), 252~J (gemma-4b), and 196~J (Qwen-3b) per query. (3) \emph{task-aware routing}---route simple instructions to low-$E_{\mathrm{tok}}$ models (Qwen: 1.02~J/tok) and hard visual reasoning to larger models; pure-text queries skip the encoder, saving $E_{\mathrm{see}}$. (4) \emph{encoder throughput optimization}---reduce $E_{\mathrm{see}} = \bar{P}\,N_{\mathrm{in}}/T_{\mathrm{prefill}}$ by decreasing image tokens (downsampling) or increasing encoder throughput (efficient vision architecture or engine, e.g., vLLM's 0.022~J/tok); keep the encoder on the GPU---the ablation shows the CPU raises energy by about 27$\times$ and latency by about 37$\times$ (Section~\ref{sec:ablation}).

\section{Limitations and Discussion}
Our conclusions are subject to several boundary conditions. First, energy is total board-level input power and includes static idle power; this cancels as a constant offset in cross-model comparisons, but absolute values depend on the power mode (Thor 120~W, Orin 50~W) and idle baseline, so absolute values across platforms are not directly comparable. Second, coverage is limited: the nine models span three architecture families but include only a few visual-encoder types (SigLIP, ViT, etc.); gemma-3-12b used early three-tier data; and the cross-engine comparison was conducted on qwen2.5-vl-3b only with GGUF Q4\_K\_M, so extrapolation to full-precision inference and other engines needs further validation. Third, the visual-prefill law is based on steady-state single-frame encoding; frame reuse can reduce actual energy when inter-frame similarity is high, but we did not perform end-to-end system-level validation on a full perception pipeline. Finally, the decode-share difference from the reference~\cite{seeing} (53--82\% vs.\ 86--97\%) stems partly from broader architectural coverage and 4-bit quantization; the qualitative laws are consistent, while quantitative values vary with setup, so cross-study comparisons should be made under identical protocols.

\section{Conclusion}
We conducted a systematic energy study of edge VLM inference on Jetson Thor and Orin, profiling nine models across three architecture families, two engines, and two hardware generations, collecting 4,508 measured records, and validating the measurement pipeline against a pure-text Qwen3 calibration standard ($R^2{=}0.9999$). We reproduced the four core findings of the reference study~\cite{seeing} on newer hardware.

Our core finding is that edge VLM energy is governed by two mutually independent energy laws. (i) The decoding energy law: $E_{\mathrm{tok}}$ is monotonically predictable from model scale ($r{=}0.970$, from 0.28~J/tok at 1B to 2.22~J/tok at 8B) and can be budgeted directly from scale and output length. (ii) The visual-prefill energy law: the visual-prefill cost is dominated by visual-encoder throughput efficiency (16$\times$ per image token), with only a weak dependence on model scale and image-token count---the 4B gemma, with only 281 image tokens (less than one-third of internvl3's 941), is paradoxically the most expensive and slowest, surpassing the 8B internvl3; onboard power is nearly constant (1.2$\times$) while throughput varies 16$\times$, so ``per-image-token visual-prefill cost $\propto 1/$encoder throughput.'' Device-mapping ablation (forcing the CPU raises energy by about 27$\times$ and latency by about 37$\times$) and cross-engine/cross-machine validation (3.4$\times$ on both Thor and Orin) confirm that the two laws are robust and first-order. The conclusion that visual-prefill energy is negligible holds only approximately when decoding is very long and the encoder is efficient; under continuous-perception workloads, the visual-prefill share rises to 92--97\% for all models, invalidating the reference conclusion~\cite{seeing}. We contribute the frame-to-answer crossover point $N^{*}$ and Joule-priced optimization strategies.

\end{document}